\documentclass[10pt,twocolumn,letterpaper]{article}

\usepackage{iccv}
\usepackage[accsupp]{axessibility}
\usepackage{times}
\usepackage{epsfig}
\usepackage{graphicx}
\usepackage{amsmath}
\usepackage{amssymb}

\usepackage{colortbl}

\usepackage{graphicx}

\usepackage{array}
\usepackage{adjustbox}
\usepackage{multirow}
\usepackage{colortbl}

\usepackage{subcaption}
\usepackage[size=small]{caption}
\usepackage{color,xcolor}
\usepackage{epsfig}
\usepackage{booktabs}

\usepackage{hhline}

\usepackage{amsmath,amsfonts,amsthm,amssymb}
\usepackage{mathtools}
\usepackage{bm}
\usepackage{nicefrac}
\usepackage{microtype}
\usepackage{algorithm, algorithmic}

\usepackage{amsmath,amsfonts,bm}

\def\eqref#1{equation~\ref{#1}}

\def\1{\bm{1}}

\def\vc{{\bm{c}}}

\def\vx{{\bm{x}}}

\DeclareMathAlphabet{\mathsfit}{\encodingdefault}{\sfdefault}{m}{sl}
\SetMathAlphabet{\mathsfit}{bold}{\encodingdefault}{\sfdefault}{bx}{n}

\newcommand{\sect}[1]{Section~\ref{#1}}

\newcommand{\fig}[1]{Figure~\ref{#1}}
\newcommand{\tbl}[1]{Table~\ref{#1}}
\newcommand{\tbls}[1]{Tables~\ref{#1}}

\DeclarePairedDelimiterX{\infdivx}[2]{(}{)}{%
  #1\;\delimsize|\delimsize|\;#2%
}

\definecolor{MyDarkBlue}{rgb}{0,0.08,1}
\definecolor{MyDarkGreen}{rgb}{0.02,0.6,0.02}
\definecolor{MyDarkRed}{rgb}{0.8,0.02,0.02}
\definecolor{MyDarkOrange}{rgb}{0.40,0.2,0.02}
\definecolor{MyPurple}{RGB}{111,0,255}
\definecolor{MyRed}{rgb}{1.0,0.0,0.0}
\definecolor{MyGold}{rgb}{0.75,0.6,0.12}
\definecolor{MyDarkgray}{rgb}{0.66, 0.66, 0.66}

\renewcommand{\subparagraph}[1]{\vspace{3pt}\noindent\emph{\underline{#1}}}
\newcommand{\myparagraph}[1]{\vspace{0.1em} \noindent \textbf{#1}}

\newcommand{\modelfull}{Neural Radiance Flow\xspace}
\newcommand{\model}{NeRFlow\xspace}

\iccvfinalcopy

\usepackage[pagebackref=true,breaklinks=true,letterpaper=true,colorlinks,bookmarks=false]{hyperref}

\def\iccvPaperID{7261} %

\ificcvfinal\fi

\begin{document}

\title{\modelfull for 4D View Synthesis and Video Processing}

\author{
\and
Yilun Du\\
MIT CSAIL\\
\and
Yinan Zhang\\
Stanford University\\
\and
Hong-Xing Yu\\
Stanford University\\
\and
\and
Joshua B. Tenenbaum\\
MIT CSAIL, BCS, CBMM\\
\and
Jiajun Wu\\
Stanford University\\
}

\maketitle

\begin{abstract}
We present a method, \modelfull (\model), to learn a 4D spatial-temporal representation of a dynamic scene from a set of RGB images. Key to our approach is the use of a neural implicit representation that learns to  capture the 3D occupancy, radiance, and dynamics of the scene. By enforcing consistency across different modalities, our representation enables multi-view rendering in diverse dynamic scenes, including water pouring, robotic interaction, and real images, outperforming state-of-the-art methods for spatial-temporal view synthesis. Our approach works even when being provided only a single monocular real video. We further demonstrate that the learned representation can serve as an implicit scene prior, enabling video processing tasks such as image super-resolution and de-noising without any additional supervision. 
\let\thefootnote\relax\footnotetext{Code at \url{https://yilundu.github.io/nerflow/}.}
\end{abstract}
\vspace{-10pt}
\section{Introduction}

We live in a rich and dynamic world, consisting of scenes that rapidly change their appearance across both time and view angle. To accurately model the world around us, we need a scene representation that captures underlying lighting, physics, and 3D structure of the scene. Such representations have diverse applications: they can enable interactive exploration in both space and time in virtual reality, the capture of realistic motions for game design, and robot perception and navigation in the environment around them.

Traditional approaches, such as those used in state-of-the-art motion capture systems, typically are specialized to specific phenomena~\cite{atcheson2008time, hawkins2005acquisition} and fail to handle complex occlusions and fine details of motion. A core difficulty is that high resolution coverage of information requires a prohibitive amount of memory. Recent work has addressed this by using a neural network as a parametrization for scene details~\cite{mildenhall2020nerf, sitzmann2019scene, Bemana2020xfields}. However, these scene representation often require a static scene and a large number of images captured from many cameras, which are not generally available in real-world scenarios. %

In this work, we aim to learn a dynamic scene representation which allows photorealistic novel view synthesis in complex dynamics, observed by only a limited number of (as few as one) cameras with known camera parameters. The key challenge is that the observations at each moment are sparse, restricting prior approaches~\cite{mildenhall2020nerf, sitzmann2019scene} from fitting a complex scene.
To address this problem, we present a novel approach, \modelfull (\model), that can effectively aggregate partial observations across time to learn a coherent spatio-temporal scene representation. We achieve this by formulating a radiance flow field, which encourages temporal consistency of appearance, density, and motion.

\begin{figure}[t]
\centering
\includegraphics[width=\linewidth]{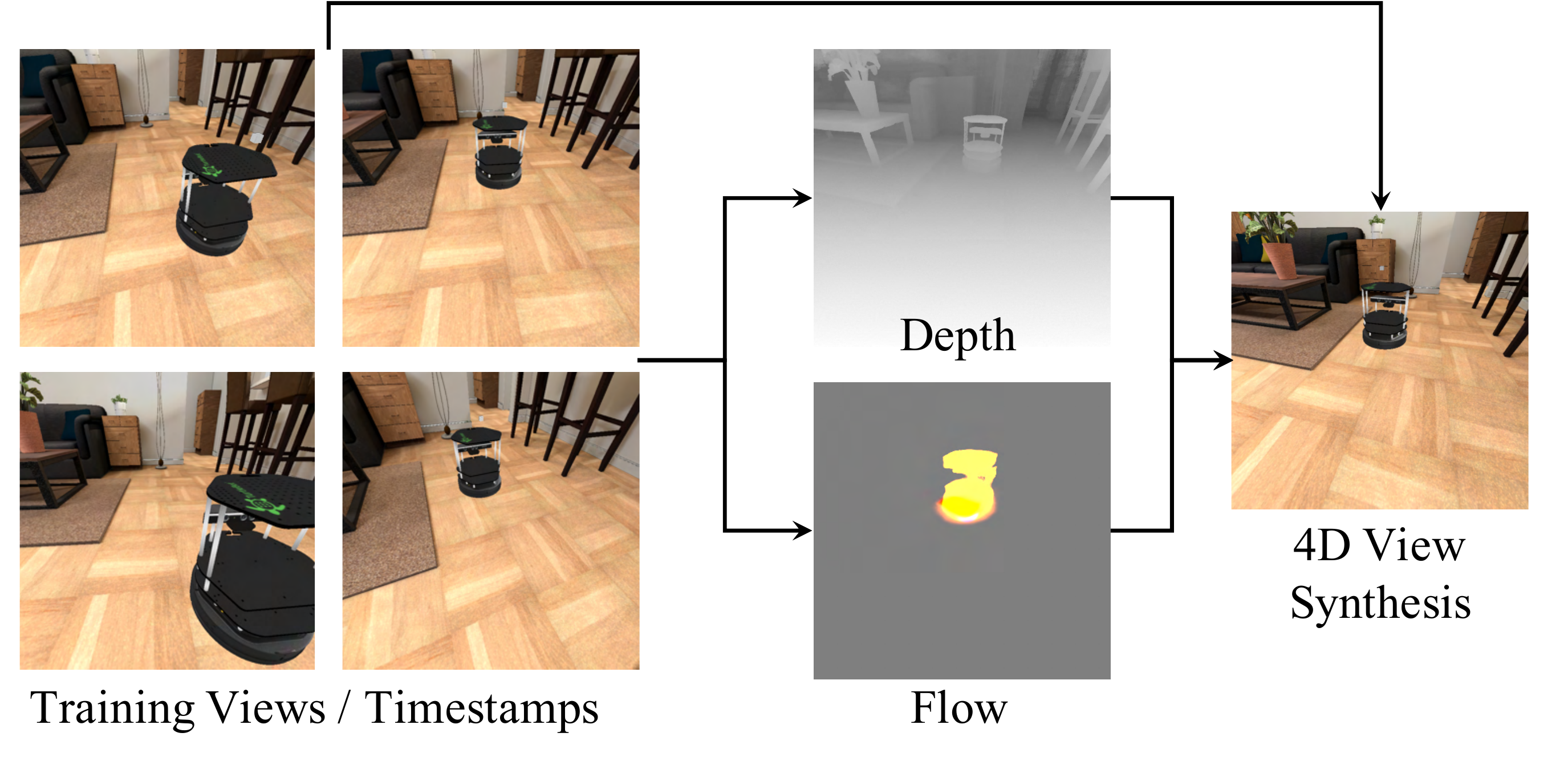}
\caption{Given a set of training images captured from different views and timestamps, \model learns a spatial-temporal representation that captures the underlying 3D structure and dynamics and, in turn, enables 4D view synthesis.}
\label{fig:teaser}
\vspace{-20pt}
\end{figure}

The radiance flow field is represented by two continuous implicit neural functions: a 6D (spatial position $x,y,z$, timestamp $t$ and viewing direction $\theta, \phi$) radiance function for appearance and density, and a 4D (spatio-temporal position $x,y,z,t$) flow function for scene dynamics. Our representation enables joint learning of both modules, which is critical given only sparse observations at each moment.  Specifically, the flow field provides temporal correspondences for spatial locations, enabling the appearance and density information captured at different moments to propagate across time. On the other hand, the radiance function describes the scene geometry that informs the flow module about how objects are moving. Our model is fully differentiable, and thus can be trained directly using gradient backpropogation. By learning 3D structure and dynamics, our model can accomplish 4D view synthesis (\fig{fig:teaser}).

To evaluate our approach, we consider several challenging setups: a pouring scene which reflects fluid dynamics, an indoor scene in which a robot walks from near to far to exhibit long-range motion with great occlusion, multiple complex real scenes with transparent objects, as well as monocular videos capturing human motions. Our approach yields high-quality 4D view synthesis and outperforms a recent state-of-the-art method~\cite{Bemana2020xfields}. In addition, we show that our method can serve as a type of dynamic scene prior, which allows video denoising and super-resolution without any additional supervision, outperforming both classical and state-of-the-art internal learning methods.

In summary, our work has three contributions. First, we present a novel method, \modelfull (\model), for learning implicit spatial-temporal scene representation. It enables novel view synthesis across both space and time. Second, we show that  our approach can be effective with very limited observations down to only one camera. We achieve this by introducing a set of temporal consistency constraints over scene appearance, density, and motion. Finally, show that our approach can serve as an implicit scene prior, outperforming classical and internal learning methods in super-resolution and image de-noising.

\section{Related Works}

\paragraph{Neural scene representations.}
Recently, neural continuous implicit fields~\cite{chen2019learning,liu2019learning,saito2019pifu,genova2019learning,xu2019disn,park2019deepsdf,oechsle2019texture,mescheder2019occupancy,sitzmann2019scene} have been developed to address the discretization issues and limited resolution of classical 3D representations such as voxel grids~\cite{brock2016generative,maturana2015voxnet,choy20163d,xie2019pix2vox,riegler2017octnet,wu20153d,wu2016learning}, point clouds~\cite{qi2017pointnet,elbaz20173d,qi2017pointnet++,fan2017point} and meshes~\cite{groueix2018papier,wang2018pixel2mesh,kato2018neural,kanazawa2018learning,liu2019soft}. Park~\etal~\cite{park2019deepsdf} proposed a neural signed distance function to represent scene geometry. Mescheder~\etal~\cite{mescheder2019occupancy} developed neural occupancy fields for scene reconstruction. However, they require groundtruth 3D supervision that can be difficult to obtain.

In order to learn neural scene representations directly from images, differentiable rendering \cite{niemeyer2020differentiable,jiang2020sdfdiff,loper2014opendr,sitzmann2019scene} is incorporated to bridge 2D observations and underlying 3D scenes. Sitzmann~\etal~\cite{sitzmann2019scene} represented scenes with continuous feature fields and propose a neural rendering layer to allow optimization with only posed images. Niemeyer~\etal~\cite{niemeyer2020differentiable} used implicit differentiation to bridge 2D images and 3D texture fields. In a recent seminal work, Mildenhall~\etal~\cite{mildenhall2020nerf} introduced a Neural Radiance Field (NeRF) that can be learned using volumetric rendering with only calibrated images. However, these works only consider static scenes.

In contrast, we aim to learn spatial-temporal dynamic scene representations with limited observations. Although it is plausible to extend existing techniques to 4D by assuming a large number of available views at each timestep, we focus on a more realistic setting in capturing dynamic events, where only a few moving cameras are available. Our setup has significance in real-world dynamic event capturing; it also poses a great challenge in aggregating sparse, partial observations across time.

\myparagraph{4D reconstruction.} Most existing works on spatial-temporal 4D reconstruction for general scenes require sufficient observations at each moment~\cite{niemeyer2019occupancy,mustafa2016temporally,mustafa2015general,leroy2017multi,ulusoy2013dynamic,oswald2014generalized}.
However, these methods need full observations at each timestamp, and they do not recover scene appearances.
Another line of work focuses on specific categories~\cite{bermano2015detailed,coskun2017long,huang2017towards,kanazawa2019learning,tung2017self,zheng20174d} such as human body and faces with template models, allowing fewer observations as input.
Using template models with deformations allows domain knowledge to be easily added and guarantees temporal coherence. Therefore this paradigm is widely adopted for particular shape domains~\cite{joo2018total,bermano2015detailed,coskun2017long,huang2017towards,kanazawa2019learning,tung2017self,zheng20174d} such as human face~\cite{bermano2015detailed}, body~\cite{kanazawa2019learning}, and hand~\cite{romero2017embodied}. However, these methods depend largely on the quality of template models and it can be costly to obtain high-quality template models beyond the popular shape domains. Unlike these methods, our \model does not make domain-specific assumptions and is able to learn from limited observations.
 
\myparagraph{Novel view synthesis.} Although synthesizing novel views in space~\cite{mildenhall2020nerf,sitzmann2019deepvoxels,nguyen2019hologan,kalantari2016learning,hedman2018deep,buehler2001unstructured,flynn2019deepview,zhou2016view,wiles2020synsin} or time (\ie, video frame interpolation)~\cite{jiang2018super,mahajan2009moving,bao2019depth,sun2018multi} is widely studied, respectively, spatial-temporal synthesis for dynamic scenes is relatively less explored~\cite{zitnick2004high,lipski2010virtual}. Recent works have extended deep learning-based novel view synthesis methods into the temporal domain, by learning a temporal warping function~\cite{lombardi2019neural,Bemana2020xfields} or synthesizing novel views frame-by-frame~\cite{bansal20204d}. Lombardi~\etal~\cite{ lombardi2019neural} modeled a scene by a neural feature volume and synthesize novel views at a given moment by sampling the volume with a temporally-specific warping function. Bemana~\etal~\cite{Bemana2020xfields} targeted at view interpolation across space and time by learning a smart warping function.
However, warping-based methods are restricted by input resolution. Our work is different from them in that we learn a continuous implicit representation that can theoretically scale to arbitrary resolution.

\myparagraph{Deep networks as prior.} Deep networks have been shown to manifest prior tendency for fitting natural images~\cite{ulyanov2018deep,van2018compressed,liu2019image} and temporally-consistent videos~\cite{lei2020blind}, even without training on large-scale datasets. This property is referred to as an implicit image/video prior. Similarly, our method can learn neural dynamic scene representations from very sparse observations. To explain such sample efficiency, we posit that our learning method per se may serve as a `dynamic implicit scene prior'. We validate it by fitting noisy and low-resolution observations, while showing good denoising and super-resolution results. Although our finding shares similar ideas with Ulyanov~\etal~\cite{ulyanov2018deep}, the prior tendency comes from our 3D rendering architecture.

\myparagraph{Concurrent Work.}
Concurrent to our work, several related works~\cite{park2020nerfies, xian2020space, li2020neural, pumarola2020dnerf, tretschk2020nonrigid} also investigate integrating temporal information for sparse time-step novel view synthesis. Separate from other works, we learn a single consistent continuous spatial-temporal radiance field that is constrained to generate consistent 4D view synthesis across both space and time. This enables direct rendering across both viewpoints and timestamps directly through the radiance field. This is not possible for other approaches which learn a discrete, timestamp-dependent deformation field~\cite{park2020nerfies, pumarola2020dnerf, tretschk2020nonrigid}. Similar to our approach, \cite{xian2020space, li2020neural} also learn a continuous spatial-temporal radiance fields, but while our approach enforces consistency across continuous time using a neural ODE~\cite{chen2018neural}, they enforce consistency only at observed timestamps. Thus, while our approach can render intermediate timestamps, \cite{xian2020space} note that interpolated renderings using their spatial-temporal radiance field are not good enough.

In addition, we further show that our approach can be applied to video processing tasks. We show that our learned radiance fields can take as input low resolution or noisy images, and can then be rendered to generate high resolution or non-noisy images.

\section{\modelfull (\model)}
\begin{figure*}[t]
\centering
\begin{minipage}[b]{.64\linewidth}
\includegraphics[width=\linewidth]{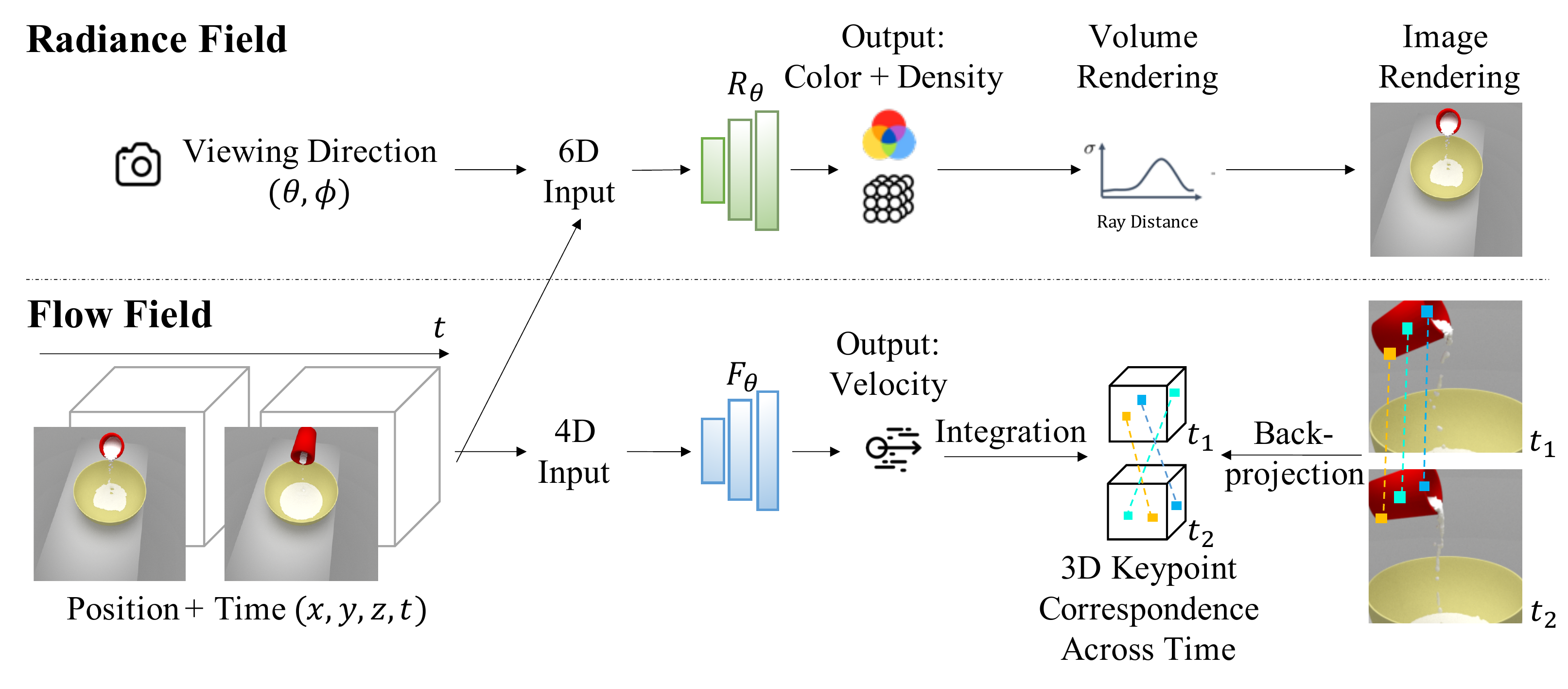}
\caption{\model consists of two separate modules, a radiance field (top) trained via neural rendering and a flow field (bottom) trained through 3D key-point correspondence. During testing, we use only the radiance field to synthesize novel images.}
\label{fig:pipeline}
\end{minipage}
\hfill
\begin{minipage}[b]{.34\linewidth}
\includegraphics[width=\linewidth]{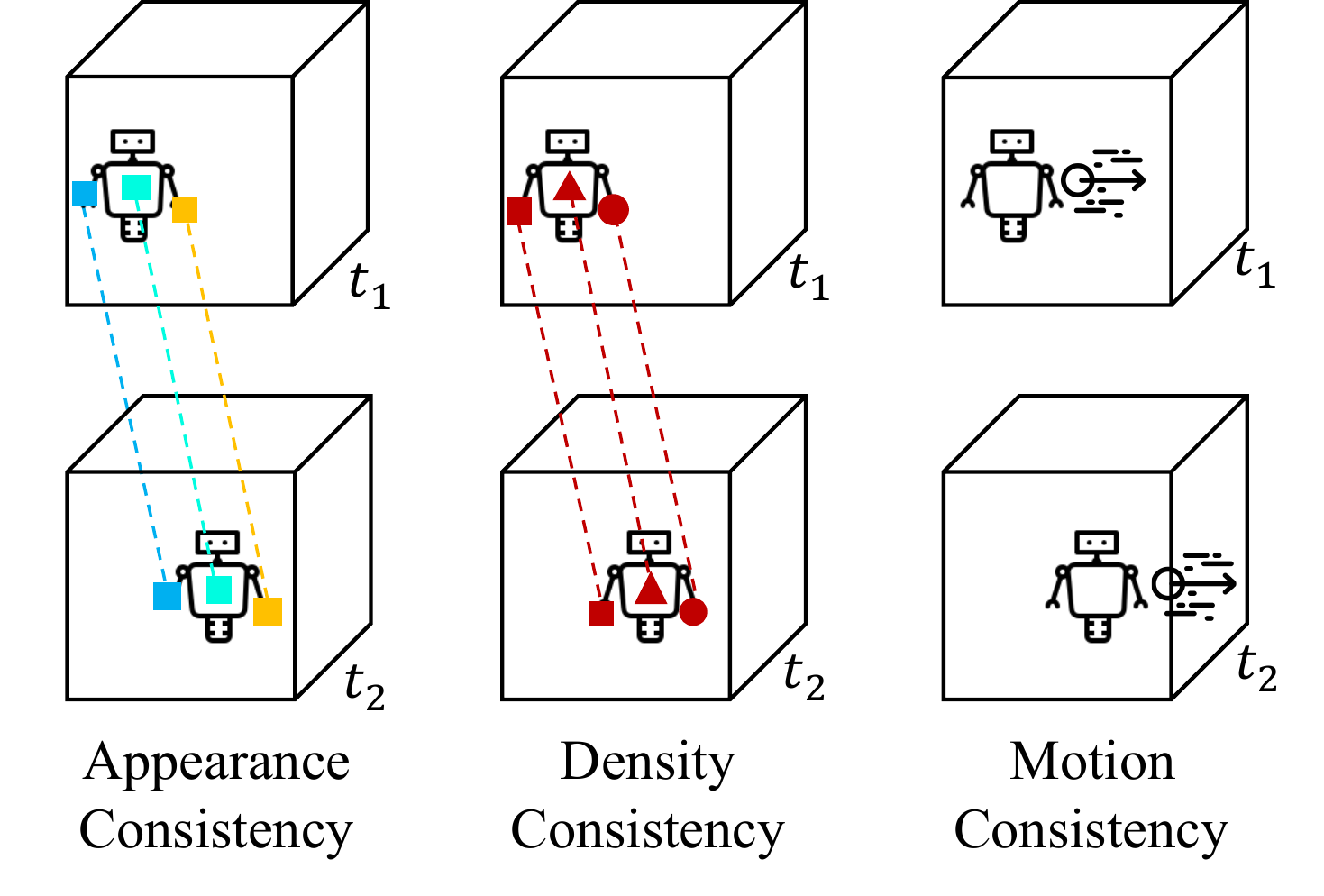}
\caption{Multiple consistencies are enforced between both radiance and flow fields during training, enabling radiance information captured at earlier timestamps to inform those of later timestamps. }
\label{fig:consistency}
\end{minipage}
\vspace{-10pt}
\end{figure*}

Our goal is to learn an implicit neural scene representation for dynamic scenes, which enables spatial-temporal novel view synthesis, given a potentially limited set of image observations and associated poses each moment. A key challenge is to effectively aggregate partial observations from different timestamps while attaining spatio-temporal coherence.

We propose \modelfull (\model) which learns scene appearance, density and motion jointly, while encouraging temporal consistency for these components. Specifically, \model internally represents scene appearance and density by a neural radiance field, and it represents scene dynamics by a flow field. These two fields interact to propagate appearance, density, and motion information observed at different moments, modulated by a set of consistency losses derived from basic physical intuitions.

We show an overview of \model in \fig{fig:pipeline}. In the following, we first describe the radiance and flow field. Next we present how the two fields are learned jointly using temporal consistency losses. Finally, we outline details about training supervision from RGB images and overall implementation.

\subsection{Radiance and Flow Fields}
\label{sect:nrf}
\model consists of two separate modules. The first module, the radiance field, takes a 6D input of position, time (encoded between -1 and 1), and view direction and outputs emitted color and density, which can be used to form an image by ray marching and volume rendering~\cite{mildenhall2020nerf}. The second module, the flow field, takes a 4D input of position and time and outputs its flow or dynamics. We describe each function in detail below. 

\myparagraph{Radiance field.} The radiance function $R_{\theta}$ is a 6-dimensional function, which takes as input the 4D location $\textbf{x} = (x, y, z, t)$ and 2D viewing direction $(\theta, \phi)$, and outputs an emitted color $\textbf{c} = (r, g, b)$ and volume density $\sigma$, representing the color and transparency of the corresponding 3D point (top of \fig{fig:pipeline}). 
Since density is view-independent, we predict the volume density $\sigma$ independently of view direction. To better aggregate cross-view visual appearance information, we also decompose the predicted color to a view-invariant diffuse part $\textbf{c}_{\text{diffuse}}$ and a view-dependent specular part $\textbf{c}_{\text{specular}}$. Since specularity is typically sparsely observed, during training we add an $\mathcal{L}_2$ regularization loss to the magnitude of $\textbf{c}_{\text{specular}}$.

\myparagraph{Flow field.} The flow function $F_{\theta}$ represents the underlying dynamics of a scene. $F_{\theta}$ takes as input the 4D location $\textbf{x} = (x, y, z, t)$. It outputs a flow field

\vspace{-5pt}{\small
\begin{equation*}
\textbf{f} = (f_x, f_y, f_z) = \left(\frac{\partial x}{\partial t}, \frac{\partial y}{\partial t}, \frac{\partial z}{\partial t}\right),
\end{equation*} 
}
representing instantaneous movement of each point in space. Through integration, this function can then be used to derive the future position of any point. In particular, given a continuous point $(x_s, y_s, z_s, t_s)$, the future position of the point at timestamp $t_g$ can be obtained through integration as
$(x_s, y_s, z_s) + \int_{t_s}^{t_g} f(x, y, z, t) dt$.

\subsection{Temporally Coherent Learning}
\label{sec:consistency}

Throughout learning, we enforce consistency in radiance and flow fields so that they interact to aggregate and propagate partially observed information across time. As shown in \fig{fig:consistency}, this internal learning process is modulated by a set of consistency losses regarding scene appearance, density, and motion, following basic physical intuitions. Since we consider sparse observations across times that may not fully cover view angles, we only enforce consistency of diffuse color $\textbf{c}_{\text{diffuse}}$ (\sect{sect:nrf}) but not specular color.

\myparagraph{Appearance consistency.}
The diffuse reflectance of an object remains constant while it is moving around. Assuming that the incidental radiance is approximately the same at the object surface \cite{ramamoorthi2001signal}, the emitted diffuse color remains constant. Such color constancy assumption is the basis for many optical flow algorithms and approximately holds especially when the motion is small. With this assumption, we develop the appearance consistency loss. 

In particular, given a randomly sampled 3D point $\textbf{x}$ at timestamp $t$, we minimize the $\mathcal{L}_2$ distance between the color of $\textbf{x}$ and that of a predicted correspondence $\textbf{x}_c$ at a future timestamp $\textbf{t}_c$ (randomly sampled between $[\textbf{t}-0.5, \textbf{t}+0.5]$)
$\mathcal{L}_\text{RGB} = \| \textbf{c}_{\text{diffuse}}(\textbf{x}) - \textbf{c}_{\text{diffuse}}(\textbf{x}_c)\|$,
where the flow function $F_{\theta}$ provides point correspondences, given by
$\textbf{x}_c = \textbf{x} + \int_{\textbf{t}}^{\textbf{t}_c} F_{\theta}(x(t)) dt$.
This appearance consistency can be seen as a way of enabling the propagation of color information gathered in earlier (or later) timestamp to that of the current timestamp. Such propagation of color representations is especially important in settings with limited dynamic cameras where visible frames are entirely disjoint from each other across time.

\myparagraph{Density consistency.}
The solidity of an object naturally remains constant while it is moving. Thus, we also enforce that density of points across time  are also consistent with respect to dynamics. Analogously to the appearance consistency, we define density consistency as
$\mathcal{L}_\text{Density} = \| \sigma(\textbf{x}) - \sigma(\textbf{x}_c)\|$.
We note that the density consistency can be particularly useful for particles like fluid, as the shape details in fluid flowing is easily missing without some form of density consistency. Our experiment in a pouring scene shows the benefit of this.

\myparagraph{Motion consistency.}
Our motion regularization is built upon two common physical intuitions: first, empty space looks static, and second, objects move smoothly in natural scenes.
For the first assumption, the static empty space should consistently have no motion. Therefore, we enforce that areas with low density must exhibit low flow. To implement this, we cast $N$ query points along a camera ray $r$, and select the first $K$ query points $q_k$ such that transmittance of the remaining camera points is greater than 0.99.  We then penalize the $\mathcal{L}_2$ magnitude of each queried point via $ \mathcal{L}_{\text{Flow}} = \| F_{\theta} (q_k) \|$.

The second assumption can be interpreted as that the overall scenes exhibit relatively low acceleration. Furthermore, a moving object (such as a walking robot) typically manifests similar flow at all points on its surface and within it body. Thus, we encourage the flow function to be smooth across both space and time, by penalizing the gradient of the flow functions at all randomly sampled points $\textbf{x}$ via
$\mathcal{L}_{\text{Acc}} = \| \nabla F_{\theta}(\textbf{x}) \|^2$.

With these consistency losses, \model can learn a spatio-temporally coherent scene representation from limited observations at different moments.

\subsection{Learning from Visual Observation}

We have introduced a temporally coherent representation of \model to model dynamic scenes. We further outline training supervision from visual observations of the scene.

\myparagraph{Volume rendering for image supervision.}
Given posed images (i.e. with camera matrices), we train our model using volumetric rendering following~\cite{mildenhall2020nerf}. In particular, let $\{(\textbf{c}_r^i, \sigma_r^i) \}$, denote the color and volume density of $N$ random samples along a camera ray $r$. We obtain a RGB value for a pixel through alpha composition:

\vspace{-10pt}{\small
\begin{equation*}
\textbf{c}_r = \sum_{i=1}^N T_r^i \alpha_r^i \textbf{c}_r^i, \quad T_r^i = \prod_{j=1}^{i-1} (1 - \alpha_r^i), \quad \alpha_r^i = 1 - \exp(-\sigma_r^i \delta_r^i),
\end{equation*}
}%
where $\delta_r^i$ denotes the sampling distance between adjacent points on a ray. We then train our radiance function to minimize the Mean Square Error (MSE) between predicted color and ground truth RGB color via
    $\mathcal{L}_{\text{Render}} = \| \textbf{c}_r - \text{RGB} \|$.

\myparagraph{Optical flow supervision.}
In addition to image supervision, we also extract optical flow correspondence using Farnback's method~\cite{farneback2003two} to serve as extra supervision for predicting better scene dynamics. We then obtain sets of 3D spatial-temporal keypoint correspondences using depth maps (ground truth for synthetic scenes and through ~\cite{Luo-VideoDepth-2020} for real video) and camera poses. Given 3D keypoint correspondences between points $\textbf{x}_s = (x_s, y_s, z_s)$ observed at timestamp $t_s$ and $\textbf{x}_g = (x_g, y_g, z_g)$ observed at timestamp $t_g$, we apply integration using an Runga-Kutta  solver ~\cite{chen2018neural} on our flow function from point $\textbf{x}_s$ to obtain a candidate keypoint $\textbf{x}^c_g$, where $\textbf{x}^c_g = \textbf{x}_s + \int_{t_s}^{t_g} F_{\theta}(x(t)) dt$. We then train our flow function to minimize the MSE between predicted and ground truth correspondences via
    $\mathcal{L}_{\text{Corr}} = \|\textbf{x}^c_g - \textbf{x}_g \|$.

\subsection{Implementation Details}

When training \model, we first use only $\mathcal{L}_{\text{Render}}$ to warm up training. We then train \model with our full loss
$ \mathcal{L}_{\text{Render}} + \mathcal{L}_{\text{Corr}} + \alpha \mathcal{L}_{\text{RGB}} + \beta \mathcal{L}_{\text{Density}} + \mathcal{L}_{\text{Flow}} + \mathcal{L}_{\text{Acc}}$, where $\alpha, \beta = 0.001$. 
We utilize the positional embedding~\cite{mildenhall2020nerf} on the inputs to the radiance function to enable the capture of high-frequency details. We omit the positional embedding in flow functions to encourage smooth flow prediction. Please see the appendix for additional training details.

\begin{table*}[t]
\footnotesize
\centering
\setlength{\tabcolsep}{3pt}
\scalebox{0.9}{
\begin{tabular}{l|cccc|cccc|cccc|cccc}
\toprule
\multirow{2}{*}{\bf Models} & \multicolumn{4}{c|}{\bf Full View} & \multicolumn{4}{c|}{\bf Stereo Views} & \multicolumn{4}{c|}{\bf Dual Views} & \multicolumn{4}{c}{\bf Sparse Timestamps} \\
\cmidrule{2-17}
& LPIPS$\downarrow$ & PSNR$\uparrow$ & SSIM$\uparrow$ & MSE$\downarrow$& LPIPS$\downarrow$ & PSNR$\uparrow$ & SSIM$\uparrow$ & MSE$\downarrow$& LPIPS$\downarrow$ & PSNR$\uparrow$ & SSIM$\uparrow$ & MSE$\downarrow$& LPIPS$\downarrow$ & PSNR$\uparrow$ & SSIM$\uparrow$ & MSE$\downarrow$ \\
\midrule
Nearest Neighbor & 0.1023 & 25.34 & 0.9858 & 0.0051 & 0.2085 & 22.89 & 0.9667 & 0.0138 & 0.1305 & 25.79 &  0.9789 & 0.0088  & 0.1237 &  24.21 & 0.9837 & 0.0061 \\
X-Fields~\cite{Bemana2020xfields} & 0.0993 & 28.83 & 0.9938 & 0.0019 & 0.1261 & 21.25 & 0.9809 & 0.0076 & 0.1190 & 20.92 & 0.9787 & 0.0082 & 0.1041 & 28.65 & 0.9933 & 0.0021 \\
NonRigid NeRF~\cite{tretschk2020nonrigid} & 0.1057 &  31.51 & 0.9968  & 0.0009  &  0.1324  & 23.38  & 0.9881  & 0.0053 &  0.1057 & 28.12  & 0.9953 & 0.0015  & - & -  & -  & - \\

\model w/o Consist. & 0.1035 & 36.30 & 0.9985 & 0.0004 & 0.1219 & 27.98 & 0.9942 & 0.0023 & 0.1021 & 31.80 & 0.9982 & 0.0006 & 0.1068 & 33.75 & 0.9980 & 0.0006 \\
\model (ours) & {\bf 0.0980} & {\bf 36.57} & {\bf 0.9990} & {\bf 0.0003} & {\bf 0.1170} & {\bf 28.29} & {\bf 0.9958} & {\bf 0.0020} & {\bf 0.0851} & {\bf 35.29} & {\bf 0.9991} & {\bf 0.0003} & {\bf 0.0949} & {\bf 35.87} & {\bf 0.9985} & {\bf 0.0004} \\
\bottomrule
\end{tabular}}
\caption{Comparison of our approach with others on the novel-view synthesis setting on the Pouring Dataset.}
\label{tbl:pouring}
\vspace{-5pt}
\end{table*}
\begin{table*}[t]
\footnotesize
\centering
\setlength{\tabcolsep}{3pt}
\scalebox{0.9}{
\begin{tabular}{l|cccc|cccc|cccc|cccc}
\toprule
\multirow{2}{*}{\bf Models} & \multicolumn{4}{c|}{\bf Full View} & \multicolumn{4}{c|}{\bf Stereo Views} & \multicolumn{4}{c|}{\bf Dual Views} & \multicolumn{4}{c}{\bf Sparse Timestamps} \\
\cmidrule{2-17}
& LPIPS$\downarrow$ & PSNR$\uparrow$ & SSIM$\uparrow$ & MSE$\downarrow$& LPIPS$\downarrow$ & PSNR$\uparrow$ & SSIM$\uparrow$ & MSE$\downarrow$& LPIPS$\downarrow$ & PSNR$\uparrow$ & SSIM$\uparrow$ & MSE$\downarrow$& LPIPS$\downarrow$ & PSNR$\uparrow$ & SSIM$\uparrow$ & MSE$\downarrow$\\
\midrule
Nearest Neighbor & 0.1945 & 17.19 & 0.8728 & 0.0219 & 0.3314 & 15.03 & 0.8501 &  0.0422 & 0.2425 & 16.86 & 0.8832 & 0.0296 & 0.2084 & 16.90 & 0.8698 & 0.0250 \\
X-Fields~\cite{Bemana2020xfields} & 0.2753 & 21.55 & 0.9410 & 0.0096 & 0.3927 & 17.75 &  0.9274& 0.0193 & 0.2587 & 19.13 & 0.9370 & 0.0142 & 0.2839 & 21.29 & 0.9378 & 0.0106 \\

NonRigid NeRF~\cite{tretschk2020nonrigid} & 0.1495 & 25.19  & 0.9616 & 0.0074 & 0.3162  & 19.43 & 0.9401  & 0.0132 & 0.2514 & 20.05 & 0.9483  & 0.0102 & - & -  & - & -  \\
\model w/o Consist.  & 0.1065 & 29.59 & 0.9846 & {\bf 0.0028} & 0.2806 & 22.47 & 0.9597 & 0.0070 & 0.2729 & 22.26 & 0.9589 & 0.0069 & 0.1130 & 25.05 & 0.9712 & 0.0072 \\

\model (ours) & {\bf 0.0984} & {\bf 30.22} & {\bf 0.9849} & 0.0029 & {\bf 0.2496} & {\bf 23.65} & {\bf 0.9690} & {\bf 0.0052} & {\bf 0.2198} & {\bf 24.84} & {\bf 0.9758} & {\bf 0.0037} & {\bf 0.1073} & {\bf 25.22} & {\bf 0.9717} & {\bf 0.0070} \\

\bottomrule
\end{tabular}
}
\caption{Comparison of our approach with others on the novel-view synthesis setting on the Gibson dataset.}
\label{tbl:gibson}
\vspace{-10pt}
\end{table*}

\section{Experiments}

We validate the performance of \model on representing dynamic scenes of pouring~\cite{schenck2016towards}, iGibson \cite{xia2020interactive}, and  real images from \cite{Bemana2020xfields,Luo-VideoDepth-2020,yoon2020novel} through multi-view rendering. We further show that our approach infers high quality depth and flow maps. Finally, we show that \model can serve as a scene prior, denoising and super-resolving videos.

\subsection{4D View Synthesis}
\label{sect:novel_view}

\paragraph{Data.} We use three datasets of dynamic scenes.

\subparagraph{Pouring}: The pouring scene contains fluid dynamics~\cite{schenck2016towards}. We render images at 400$\times$400 pixels. We utilize a training set size of 1,000 images and test set of 100 images.

\subparagraph{Gibson}: The Gibson scene has a robot walking a long distance. We render images on the iGibson environment~\cite{xia2020interactive}, using the \texttt{\small Rs\_interactive} scene 
    with a robot TurtleBot moving linearly on the floor. Each image is rendered at 800$\times$800 pixels. We use a training set size of 300 images and test set size of 100 images. 
    
\subparagraph{Real Images}: Our real image datasets consists of two sources. The first are two real dynamic scenes from~\cite{Bemana2020xfields}, named Ice and Vase, where Ice contains transparent objects and Vase is a complex indoor scene. The second is monocular real world videos from~\cite{Luo-VideoDepth-2020} and ~\cite{yoon2020novel}. To evaluate on two real dynamic scenes from ~\cite{Bemana2020xfields}, we split 90\% of provided images as the training set and the remaining 10\% as the test set, and use COLMAP~\cite{schoenberger2016sfm} to obtain poses for all images.  

\myparagraph{Metrics.} To measure the performance of our approach, we report novel view synthesis performance using LPIPS~\cite{Zhang2018Unreasonable}, PSNR, SSIM~\cite{Wang2004Image}, and MSE. 

\myparagraph{Baselines.} We compare with four baselines. The first is the nearest neighbor baseline from Open4D~\cite{bansal20204d} (using VGG feature distance). The second is a recent state-of-the-art method, X-Fields~\cite{Bemana2020xfields}, which relies on warping existing training images to synthesize novel views. The third is a concurrent work, NonRigid NeRF~\cite{tretschk2020nonrigid}, using the author's provided codebase. We also compare with ablations.

\begin{figure}[t]
\centering
\includegraphics[width=\linewidth]{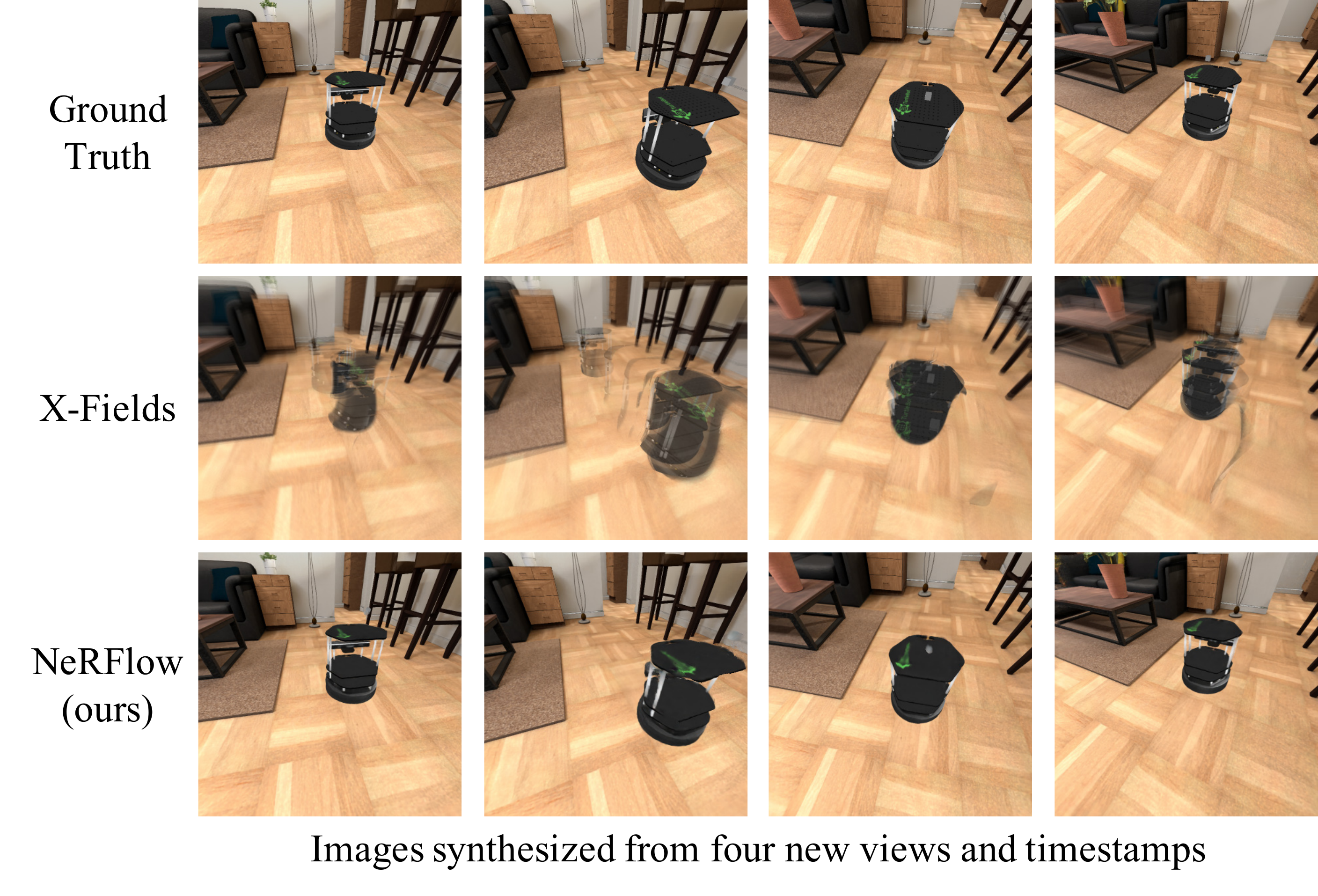}
\vspace{-15pt}
\caption{Results on Gibson in the Full View setting.
}
\vspace{-10pt}
\label{fig:pouring_gibson_qual}
\end{figure}

\begin{figure}
\centering
\includegraphics[width=.9\linewidth]{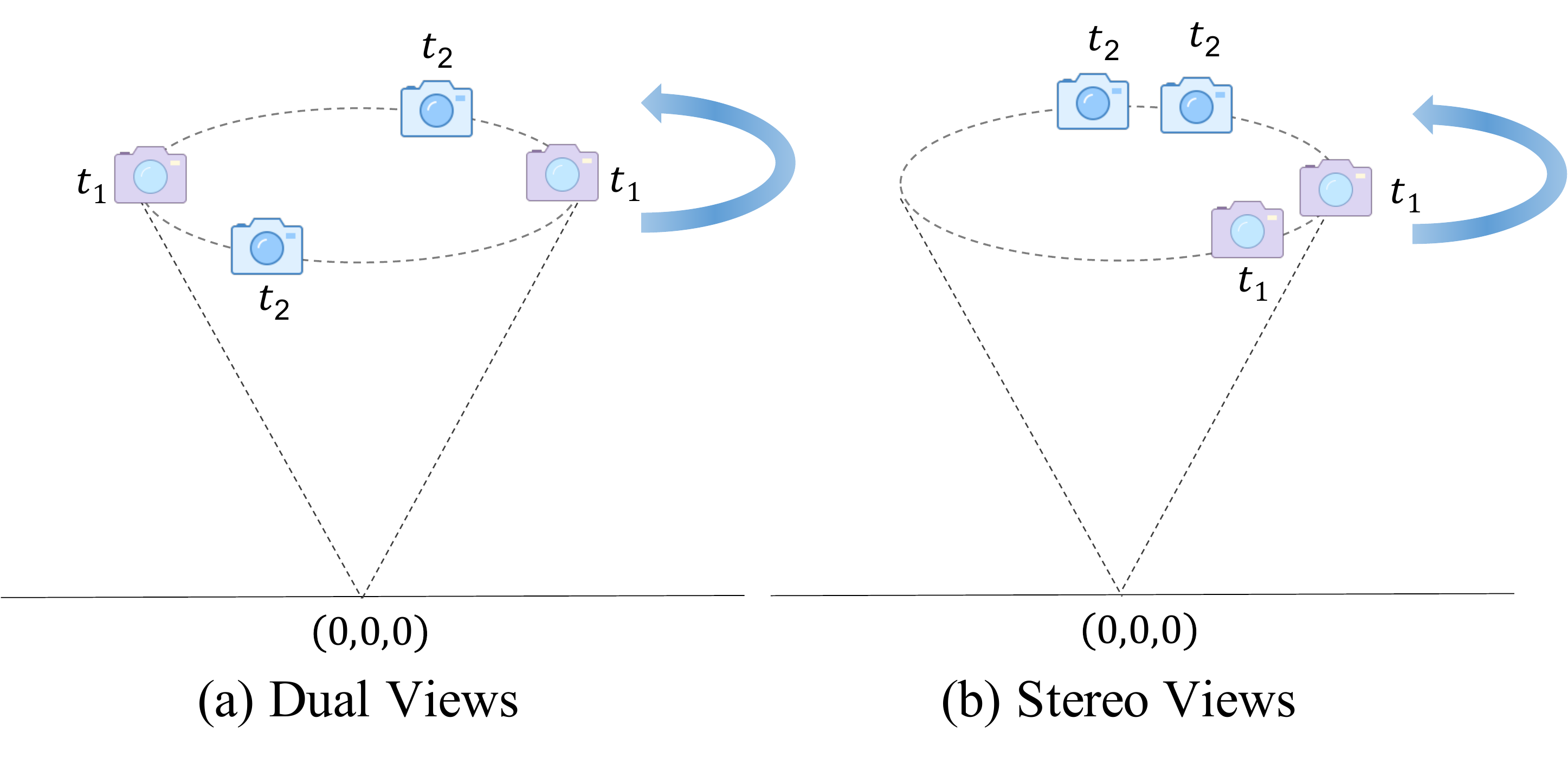}
\caption{Illustration of cameras in the limited views setting.}
\label{fig:limited}
\vspace{-15pt}
\end{figure}

\begin{figure*}[t!]
\centering
\includegraphics[width=\linewidth]{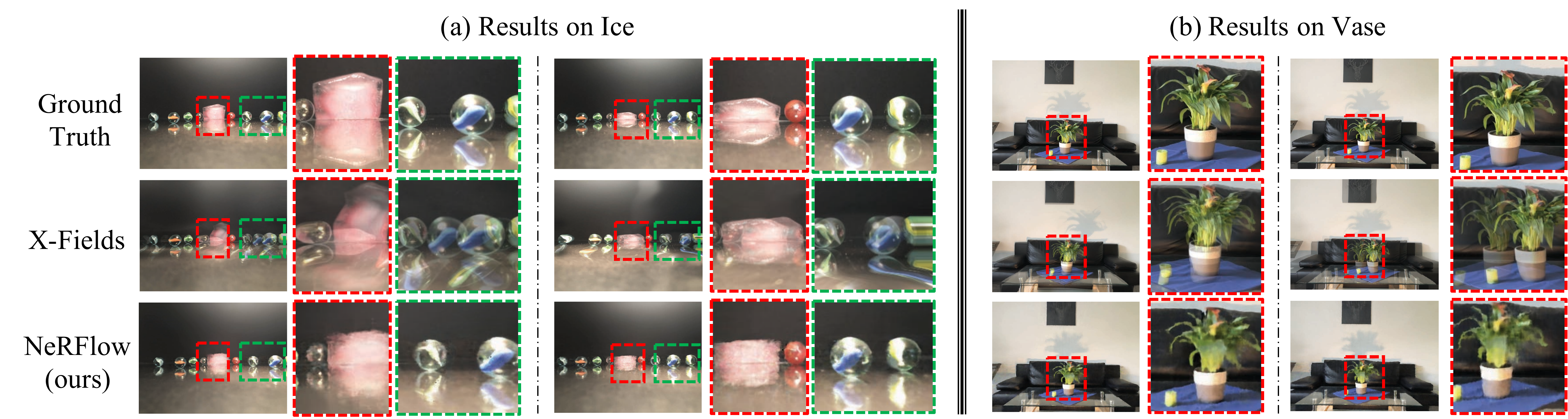}  

\vspace{3pt}
\footnotesize
\setlength{\tabcolsep}{4pt}
\hspace{65pt}
\begin{tabular}{lcccc}
\toprule
\bf Model  & LPIPS$\downarrow$ & PSNR$\uparrow$ & SSIM$\uparrow$ & MSE$\downarrow$ \\
\midrule
X-Fields~\cite{Bemana2020xfields} & 0.2271 & 18.69 & 0.9347 & 0.0140 \\
\model (Ours) & \textbf{0.2031} & \textbf{29.04} & \textbf{0.9922} & \textbf{0.0012} \\
\bottomrule
\end{tabular}
\hfill
\setlength{\tabcolsep}{4pt}
\begin{tabular}{lcccc}
\toprule
\bf Model  & LPIPS$\downarrow$ & PSNR$\uparrow$ & SSIM$\uparrow$ & MSE$\downarrow$ \\
\midrule
X-Fields~\cite{Bemana2020xfields} &  0.2151 & 19.92 & 0.9259 & 0.0105 \\
\model (Ours) & \textbf{0.1972} & \textbf{28.91} & \textbf{0.9851} & \textbf{0.0013} \\ 
\bottomrule
\end{tabular}
\caption{Novel view synthesis results on Ice (a) and Vase (b) from the sparse image datasets used by X-Fields~\cite{Bemana2020xfields}.}
\label{fig:xfield}
\vspace{-15pt}
\end{figure*}

\myparagraph{Results on synthetic images.} On synthetic images, we present a systematic analysis in three different settings for benchmarking different methods: 1) Full View, where multi-view training images are drawn uniformly across time; 2) Stereo Views or Dual Views, where training images are captured by two moving cameras; 3) Sparse Timestamps, where training images are drawn from a fixed, sparse subset of all timestamps in the scene. 

\subparagraph{`Full View' results.} In the Full View setting, for the Pouring dataset, we sample cameras poses randomly in the upper hemisphere; for the Gibson dataset, we sample cameras from a set of forward facing scenes. 

\tbls{tbl:pouring} and \ref{tbl:gibson} (Full View) include quantitative results on the Pouring and Gibson datasets, respectively. \model outperforms baselines in all metrics. We find that on Pouring, where modeling fluid dynamics is difficult, \model is able to capture the fluid splatter pattern and dynamics. On Gibson, which exhibits long range motion and occlusion (\fig{fig:pouring_gibson_qual}), \model is able to handle of occlusions of robot. %

\begin{figure*}
    \centering
    \vspace{-10pt}
    \includegraphics[width=\linewidth]{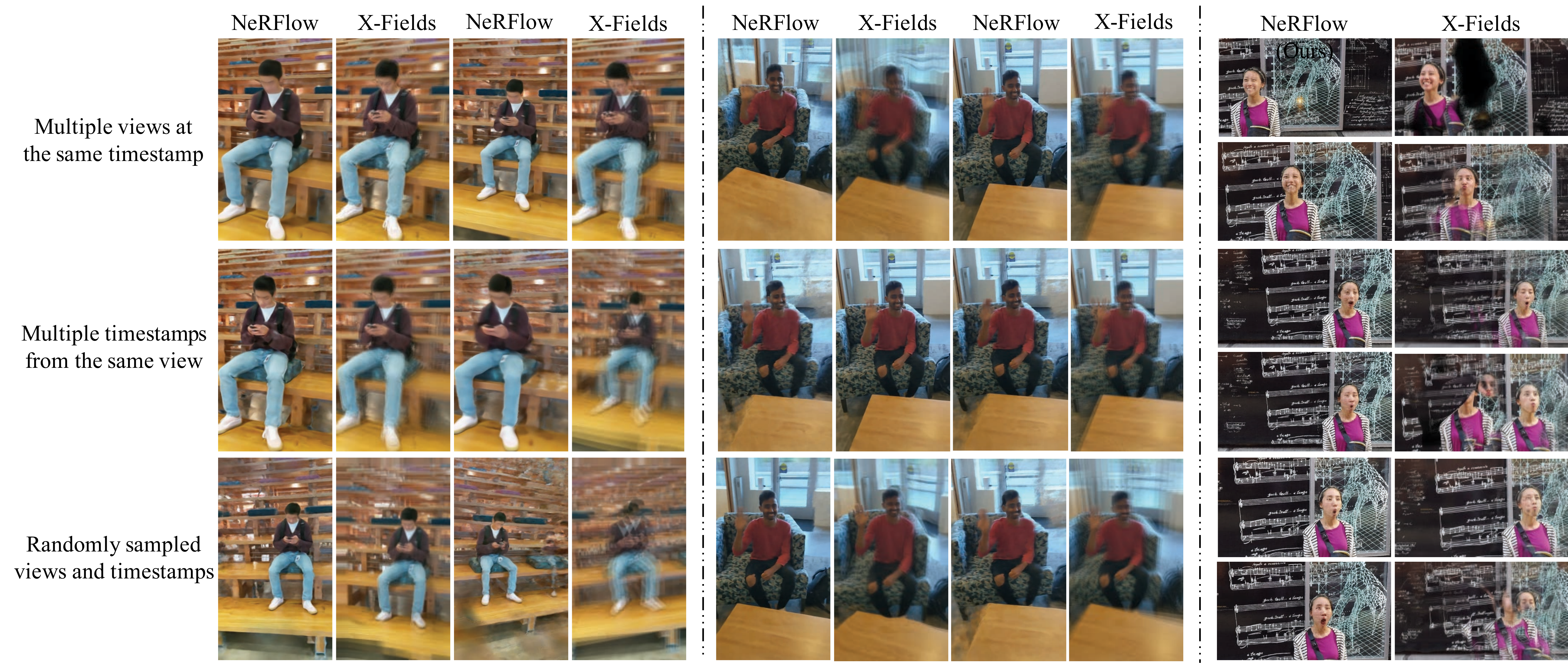}
    \caption{Comparison of 4D synthesis (novel view and timestamp) with X-Fields on real monocular video from \cite{yoon2020novel, Luo-VideoDepth-2020}.}
    \label{fig:video_real}
    \vspace{-15pt}
\end{figure*}

\subparagraph{`Stereo Views' and `Dual Views' results.} In the Stereo Views setting, training images are captured by two nearby cameras that are rotating together around a circle over time. In the Dual Views setting, training images are captured by two diametrically opposite cameras that are rotating together around a circle over time. We illustrate both settings in \fig{fig:limited}. We test image synthesis from random views on any location of the circle across any time.  To accomplish this task well, a model must learn to integrate the radiance information captured across different timestamps. 

We report the results in \tbls{tbl:pouring} and \ref{tbl:gibson} (Stereo Views and Dual Views). Our model again outperforms all baselines. In this setting, we find that consistency enables our approach to do significantly better quantitatively, which we will elaborate later in our ablation studies in \sect{sec:ablation}.

\subparagraph{`Sparse Timestamps' results.} We further consider the case where training images are drawn from a fixed, sparse subset of all timestamps in the scene. In particular, we train models with 1 of every 10 timestamps on Pouring, and 1 out of every 5 on Gibson. During testing, the model needs to render at arbitrary timestamps. Such a task tests the temporal interpolation capability of our model, useful for applications such as slow motion generation and frame rate up-conversion.

We report quantitative results in \tbls{tbl:pouring} and \ref{tbl:gibson} (Sparse Timestamps). NonRigid NeRF is not applicable to this setting, as it learns per timestamp latents for each timestamp in the scene.
Again, \model performs well and consistency boosts the rendering performance. Consistency constrains radiance fields to change smoothly with respect to time, enabling smooth renderings of intermediate timestamps. Due to space constraints, qualitative results and additional analyses can be found in the supplementary material. %

\myparagraph{Results on real images.}

We further evaluate our approach on real images: we perform novel view synthesis on the image dataset in X-Fields~\cite{Bemana2020xfields}, and 4D view synthesis on monocular real video datasets from \cite{Luo-VideoDepth-2020, yoon2020novel}.  \fig{fig:xfield} shows the novel view synthesis results on the X-Fields dataset. We find that \model captures transparency and various lighting effects in real images, while X-Fields struggles with ghosting. 

We also show 4D view synthesis results in \fig{fig:video_real} on monocular video datasets from \cite{Luo-VideoDepth-2020, yoon2020novel}. We visualize three different sets of results on 4D video synthesis: (1) multiple views at the same timestamp; (2) multiple timestamps from the same view; (3) randomly sampled views and timestamps.  In all cases, the combination of the timestamp and the view are not in the training set. \model consistently delivers better results. Note that for the leftmost video, X-Fields appears to get incorrect poses of rendering due to a dominance of temporal warping. Additional monocular video results are in the supplementary material. %

\subsection{Analysis and Visualization}
\label{sec:ablation}

\begin{figure}[t]
\centering
\includegraphics[width=\linewidth]{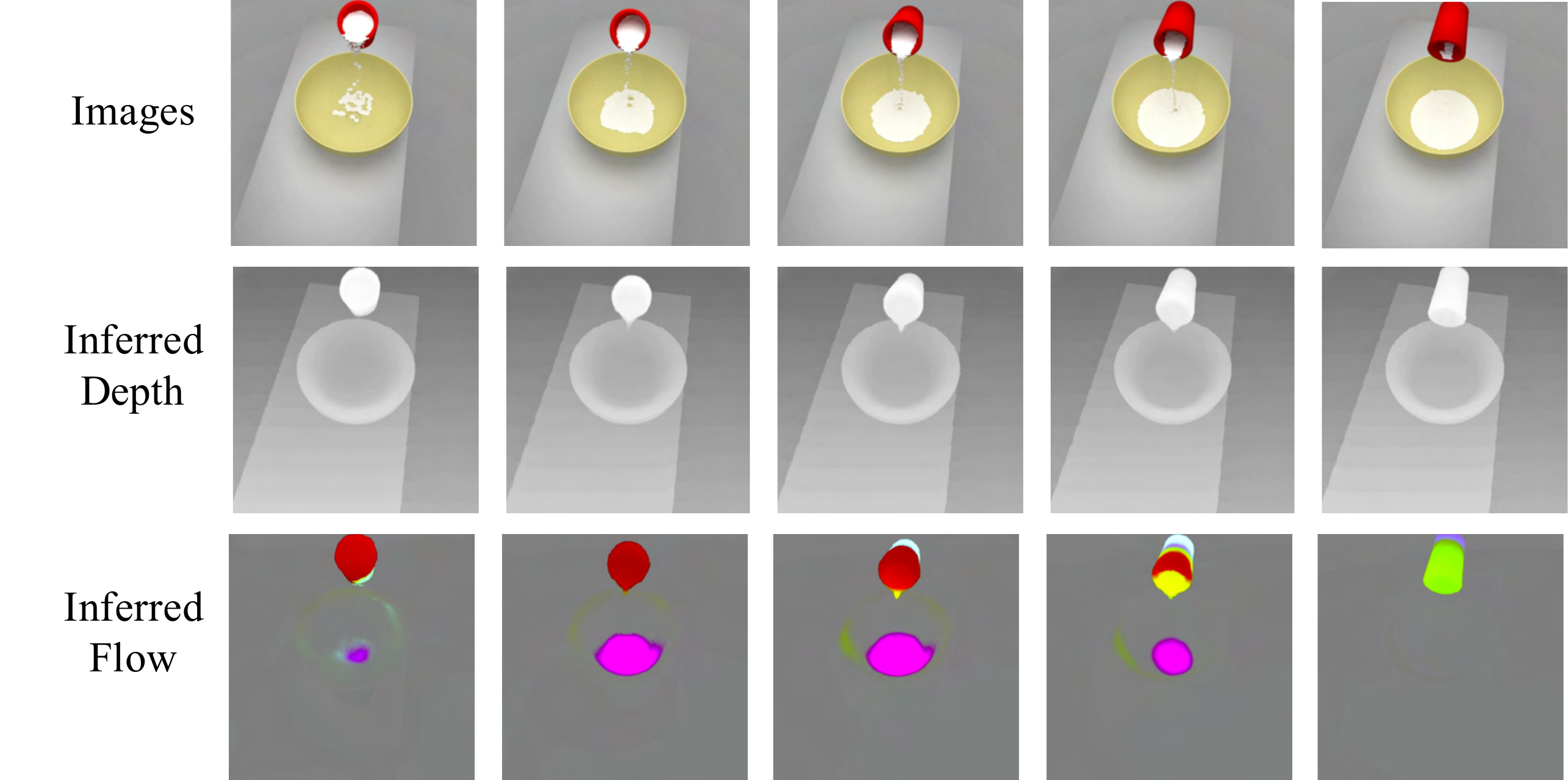}
\caption{Visualization of estimated depth and flow (with x, y, z directions of flow represented as RGB coordinates respectively).}
\label{fig:3d_flow_pouring}
\vspace{-10pt}
\end{figure}

We next analyze \model to visualize its learned depth and flow maps, and investigate how consistency losses contribute to the learning of such representations and to the final results. We run our analyses on Pouring, as its simplicity leads to most interpretable results.

\myparagraph{Visualization of depth and flow maps.}  In \fig{fig:3d_flow_pouring}, we visualize inferred depth and flow fields. We find that the inferred flow field captures the dynamics of pouring, including the flow of liquid as well as the movement of the cup. We quantitatively compare the depth estimation accuracy of \model with and without all consistency losses in \tbl{tbl:quant_depth} in terms of MSE. By enforcing geometric constancy across time, we find that our consistency loss improves depth estimates from \model, especially in limited camera settings.

\begin{table}[t]
\footnotesize
\label{tbl:depth}
\setlength{\tabcolsep}{3pt}
\centering
\begin{tabular}{lccc}
\toprule
\bf Models & \bf Full View & \bf Stereo Views & \bf Dual Views \\
\midrule
\model w/o Consistency & 0.3747  & 0.4433  & 0.4003\\
\model (ours) & \textbf{0.3692} & \textbf{0.2701} &  \textbf{0.2675} \\
\bottomrule
\end{tabular}
\caption{Evaluation of depth estimation of \model with or without physical constraints. We report MSE error with ground truth depth.}
\label{tbl:quant_depth}
\vspace{-15pt}
\end{table}

\myparagraph{Ablation study of consistency losses.}
Our consistency loss can be seen as a way to explicitly enforce the separation of static and dynamic components.  When scene flow is predicted to be near 0, static temporal consistency of appearance/density is enforced while non-zero scene flow propagates radiance information for dynamic modeling. We enforce zero scene flow at static locations through noisy optical flow and \emph{motion consistency} ($L_{flow}$ and $L_{acc}$).  

\begin{figure}[t]
\centering
\includegraphics[width=\linewidth]{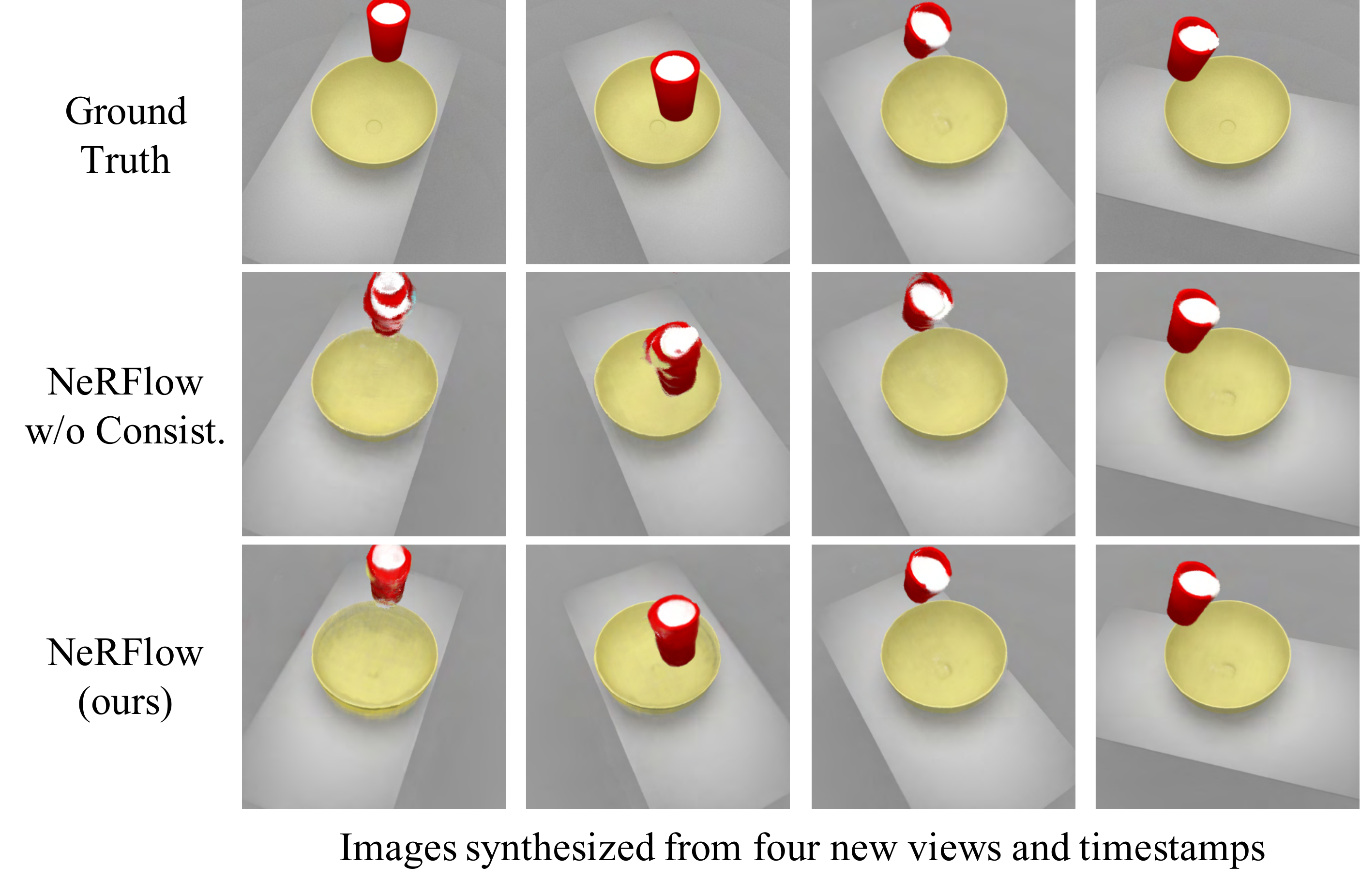}
\footnotesize
\begin{tabular}{lcccc}
\toprule
\bf Models & LPIPS$\downarrow$ & PSNR$\uparrow$ & SSIM$\uparrow$ & MSE$\downarrow$\\
\midrule
  w/o Optical Flow & 0.1390  &  27.81 & 0.9932 &  0.0023 \\
  w/o Consist. & 0.1219  &  27.98 & 0.9942 &  0.0023 \\
  w/o Motion Consist. &0.1372 & 27.85 & 0.9935 & 0.0023  \\
  w/o Dynamic Modeling &0.1317 & 28.09  & 0.9938 & 0.0022 \\
 \model (Full) &  \textbf{0.1170} & \textbf{28.29}  &  \textbf{0.9958} & \textbf{0.0020} \\
\bottomrule
\end{tabular}
\caption{\label{fig:stereo_pouring} Ablation study of different consistency losses on the Stereo Views setting of Pouring. Consistency regularization ensures reasonable renderings in extrapolated viewpoints. We also visualize the results of our model with or without consistency regularization.}
\vspace{-5pt}
\end{figure}

We now ablate the effect of this separation by considering two variants: (1) reducing static separation by removing motion consistency (w/o Motion Consist.); (2) removing the dynamic modeling by only enforcing consistency on points with flow below the threshold 0.01 (w/o Dynamics Modeling). \fig{fig:stereo_pouring} shows quantitative results of these models, in addition to our full and an ablated model without any consistency, on the Stereo Views setting of Pouring. Qualitatively, consistency enables more effective propagation of information across time. It makes renderings exhibit more consistent fluid placement. %
The supplementary material includes additional qualitative results of each variant of our ablation, which demonstrate that reducing static supervision leads to poor static structures and removing dynamic modelings leads to poor modeling of dynamic regions.

\subsection{Video Processing}

Given a set of images capturing a dynamic scene, \model learns to represent the underlying 3D structure and its evolution through time. This scene description can be seen as a scene prior. By utilizing volumetric rendering on our scene description, we accomplish additional video processing tasks such as video denoising and super-resolution. 

\myparagraph{Datasets.} We evaluate our approach on the tasks of video denoising and image super-resolution. To test denoising, we train our model on 1,000 pouring images of the same scene with a resolution of 400$\times$400, rendered with a 2 ray-casts (compared with 128 rays used in \sect{sect:novel_view}) in Blender, and test the difference between the rendered images and the ground truth images obtained from Blender using 128 ray-casts. We also evaluate our approach on denoising a monocular real video (Ayush) from \cite{Luo-VideoDepth-2020}, where we corrupt input frames with Gaussian noise with standard deviation of 25. 
To test super-resolution, we train our model on 1,000 pouring images of the same scene with a resolution of 64$\times$64 and test rendering of images of size 200$\times$200. 

\myparagraph{Baselines.} We compare with the very recent, state-of-the-art internal learning method, Blind Video Prior~\cite{lei2020blind}, which uses a learned network to approximate a task mapping. During training, we supervise the Blind Video Prior on denoising and super-resolution using the outputs of the classical algorithms: Non-Local Means~\cite{buades2005nonlocal} for denoising and bi-cubic interpolation for super-resolution. We also compare with these classical algorithms directly. 

\begin{figure}[t]
\centering
\includegraphics[width=\linewidth]{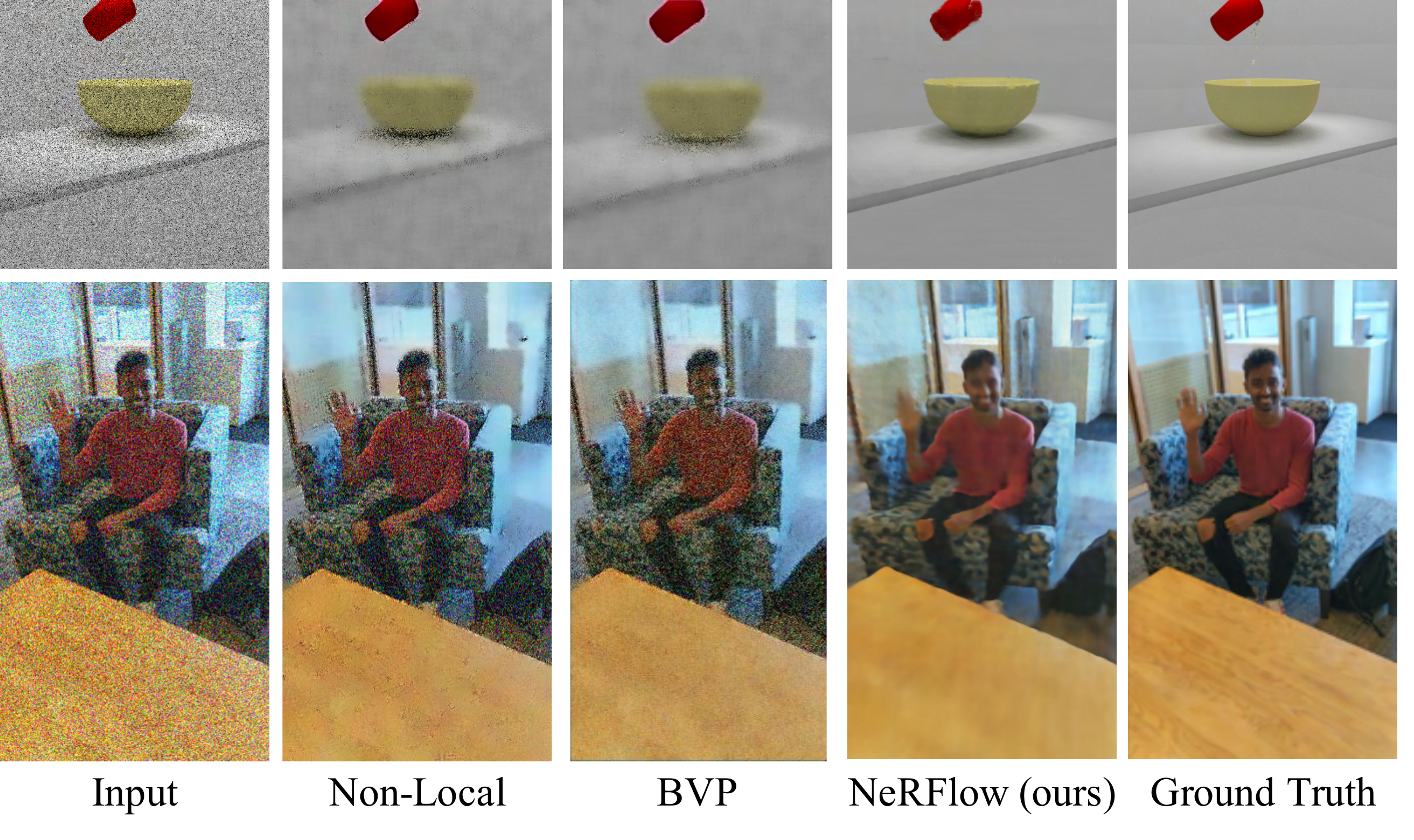}
\footnotesize
\centering
\setlength{\tabcolsep}{3pt}
\begin{tabular}{llcccc}
\toprule
\bf Data & \bf Models & LPIPS$\downarrow$ & PSNR$\uparrow$ & SSIM$\uparrow$ & MSE$\downarrow$\\
\midrule
\multirow{3}{*}{Pouring} & Non-Local Means~\cite{buades2005nonlocal} & 0.4662 &  24.91 & 0.9263 & 0.0032 \\
 & Blind Video Prior~\cite{lei2020blind} & 0.5572 & 18.24 &0.8891 & 0.0151\\
 & \model (ours) & \textbf{0.3556} & \textbf{28.46} & \textbf{0.9837} & \textbf{0.0014} \\
\midrule
\multirow{3}{*}{Ayush~\cite{Luo-VideoDepth-2020}} &  Non-Local Means~\cite{buades2005nonlocal} & 0.3051  & 23.49  &  0.9856 & 0.0046  \\
 & Blind Video Prior~\cite{lei2020blind} & 0.2707 & 21.67 & 0.9797 & 0.0070 \\
 & \model (ours) & \textbf{0.1372}  & \textbf{27.71} & \textbf{0.9949} & \textbf{0.0018} \\
\bottomrule
\end{tabular}
\caption{Results of \model, Blind Video Prior~\cite{lei2020blind}, and Non-Local Means~\cite{buades2005nonlocal} on the task of denoising.}
\label{fig:denoising}
\vspace{-15pt}
\end{figure}

\myparagraph{Video denoising.}
\fig{fig:denoising} shows the results on denoising. \model achieves more realistic images than the baselines and a lower reconstruction error. %
By accumulating radiance information over input images, our representation learns to remove most image noises. On the real monocular video, \model also obtains more realistic images than our baselines (as in LPIPS) and achieves a lower MSE.

\begin{table}[t]
\footnotesize
\centering
\begin{tabular}{lcccc}
\toprule
 \bf Model & LPIPS$\downarrow$ & PSNR$\uparrow$ & SSIM$\uparrow$ & MSE$\downarrow$\\
\midrule
Bicubic Interpolation & 0.1427 & 30.27  & 0.9961  & 0.0012 \\
Blind Video Prior~\cite{lei2020blind} &  0.1870 & 30.58 & \textbf{0.9963} & \textbf{0.0009} \\
\model (ours)  & \textbf{0.0903} & \textbf{30.67} & \textbf{0.9963} & \textbf{0.0009} \\
\bottomrule
\end{tabular}
\caption{Results of \model, Blind Video Prior~\cite{lei2020blind}, and bi-cubic interpolation on the task of image super-resolution.}
\label{tbl:superres}
\vspace{-15pt}
\end{table}

\myparagraph{Video super-resolution.}
We finally evaluate our approach on image super-resolution with our baselines in \tbl{tbl:superres}. We find that in this setting our approach again achieves more realistic images than our baselines (as determined by LPIPS), with the Blind Video Prior achieving comparable image MSE. When rendering higher resolution images from our radiance function, representations in \model have accumulated radiance information across different input images, and are capable of rendering higher resolution details, despite being only trained on low resolution images. The supplementary material includes qualitative results. %

\vspace{-0.15cm}
\section{Discussion}
We have presented \model, a method that learns a powerful spatial-temporal representation of a dynamic scene. We have shown that \model can be used for 4D view synthesis from limited cameras (\eg, monocular videos) on multiple datasets. We have also shown that \model can serve as a learned scene prior, which can be applied to video processing tasks such as video de-noising and super-resolution.

\myparagraph{Limitations:} Representing a dynamic 3D scene for view synthesis from limited image observations poses great challenges besides information aggregation. Our approach does not explicitly address the ambiguities in both 3D geometry and dynamic regions.  Such ambiguity leads to difficulty in modeling complex real scenes and in preserving static backgrounds over time. We envision addressing these two challenges can greatly improve our approach, e.g., explicitly separating static background and dynamic foregrounds to determine which regions should have non-zero flows, and leveraging dense depth maps to resolve geometry ambiguity.

\myparagraph{Acknowledgements:} Yilun Du is funded by an NSF graduate fellowship. This work is in part supported by ONR MURI N00014-18-1-2846, IBM Thomas J. Watson Research Center CW3031624, Samsung Global Research Outreach (GRO) program, Amazon, Autodesk, and Qualcomm.

{\small
\bibliographystyle{ieee_fullname}
\bibliography{egbib, reference}
}

\newcommand{\appendixhead}%
{\centering\textbf{\Large Appendix: Neural Radiance Flow for 4D View Synthesis and Video Processing}
\vspace{0.25in}}

\twocolumn[\appendixhead]
\maketitle
\appendix

We first present additional qualitative visualizations in \sect{sect:visualization} on both real and synthetic images.
We further describe training details in \sect{sect:training_details} and provide pseudocode for the overall algorithm. We recommend reviewers to see attached videos for qualitative visualizations of our approach.

\section{Additional Visualizations}
\label{sect:visualization}

\begin{figure}[h]
\centering
\includegraphics[width=\linewidth]{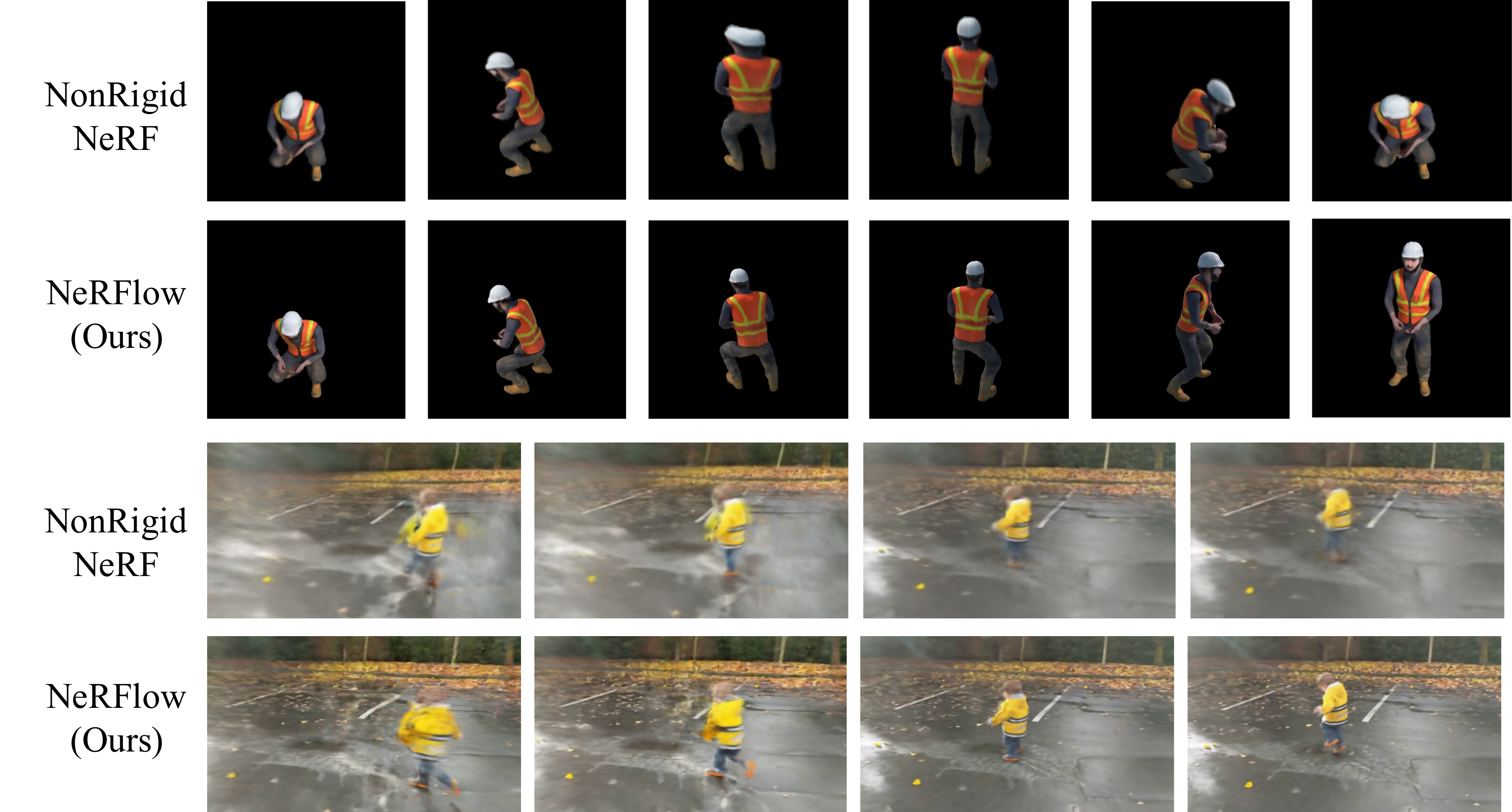}
\caption{\small Novel view renderings on dynamic scenes. Note that NonRigid NeRF collapsed to an earlier frame in the first row's last image, and our NeRFlow performs better. See the video for details.}
\label{fig:more_dynamic}
\vspace{-10pt}
\end{figure}

\subparagraph{Real Image View Synthesis}: We illustrate novel view renderings in scenes with greater degrees of underlying dynamics from ~\cite{li2020neural} and ~\cite{pumarola2020dnerf} in \fig{fig:more_dynamic}. Compared with NonRigid NeRF, our renderings are sharper and capture dynamics more accurately. 

\begin{figure}[h]
\centering
\includegraphics[width=\linewidth]{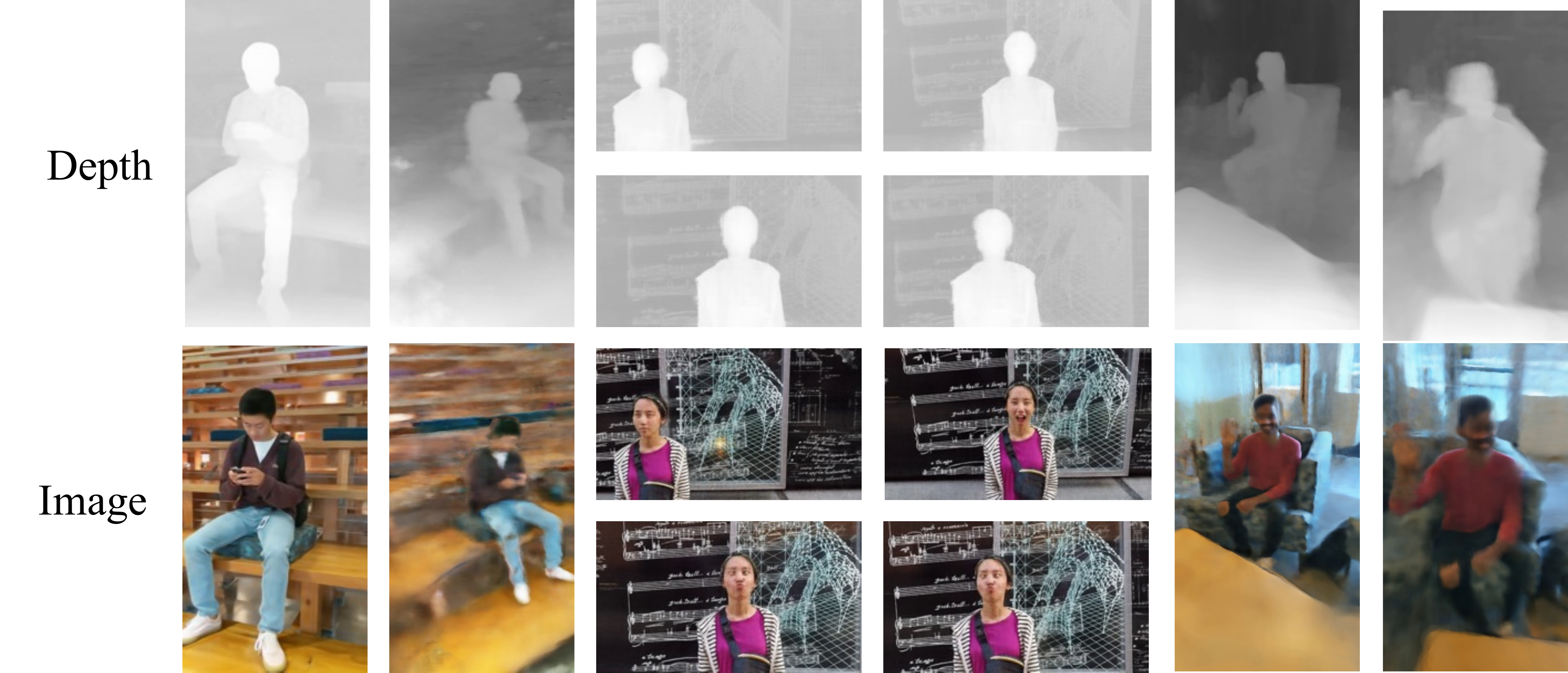}
\caption{\small Illustrations of depth predicted by NeRFlow. NeRFlow is able to reliably capture the depth of dynamic scenes.}
\label{fig:rgb_depth}
\vspace{-10pt}
\end{figure}

\subparagraph{Real Depth Maps}: We visualize depth images from a variety of real images in \fig{fig:rgb_depth}. We find that NeRFlow is able to reliably infer the depth of a variety of images.

\begin{figure*}[t!]
\centering
\includegraphics[width=\linewidth]{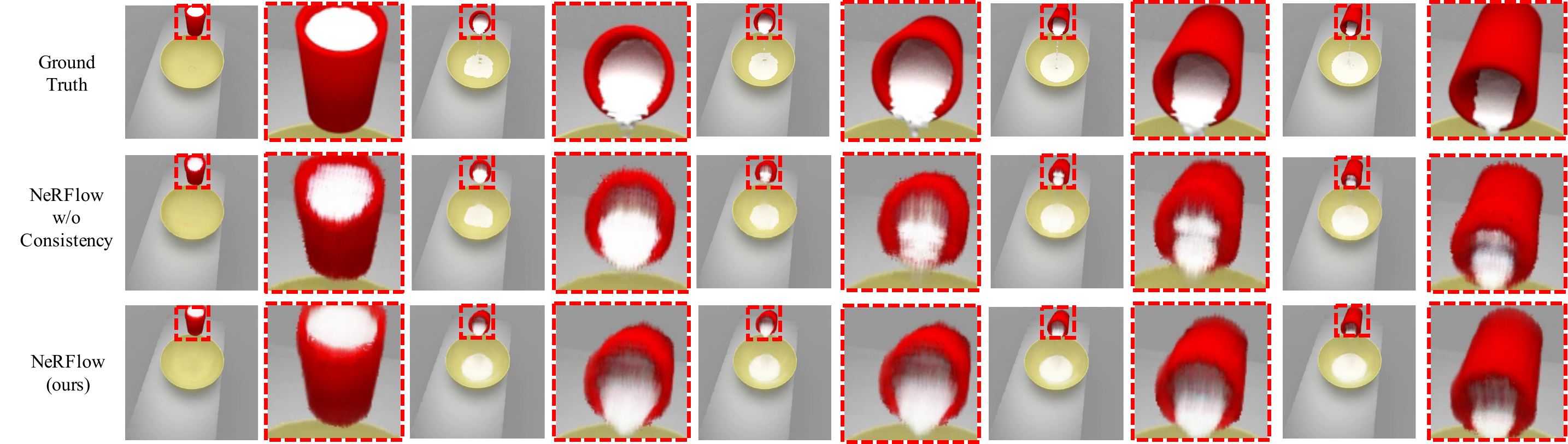}
\caption{Illustration of rendering results on intermediate timestamps when models are trained with a sparse number of timestamps. 
Consistency makes pouring volume significantly more stable.  
}
\label{fig:pouring_consistency}
\vspace{-10pt}
\end{figure*}

\subparagraph{Image Interpolations}: We provide visualizations of temporal interpolations of \model. \model is trained with 1 in 5 frames in the pouring scene, and rendered across all frames in a scene.  With consistency, we find that we are able to consistently model drops of liquid volume throughout the duration of the pouring animation (\fig{fig:pouring_consistency}).

\begin{figure}[h]
\centering
\includegraphics[width=\linewidth]{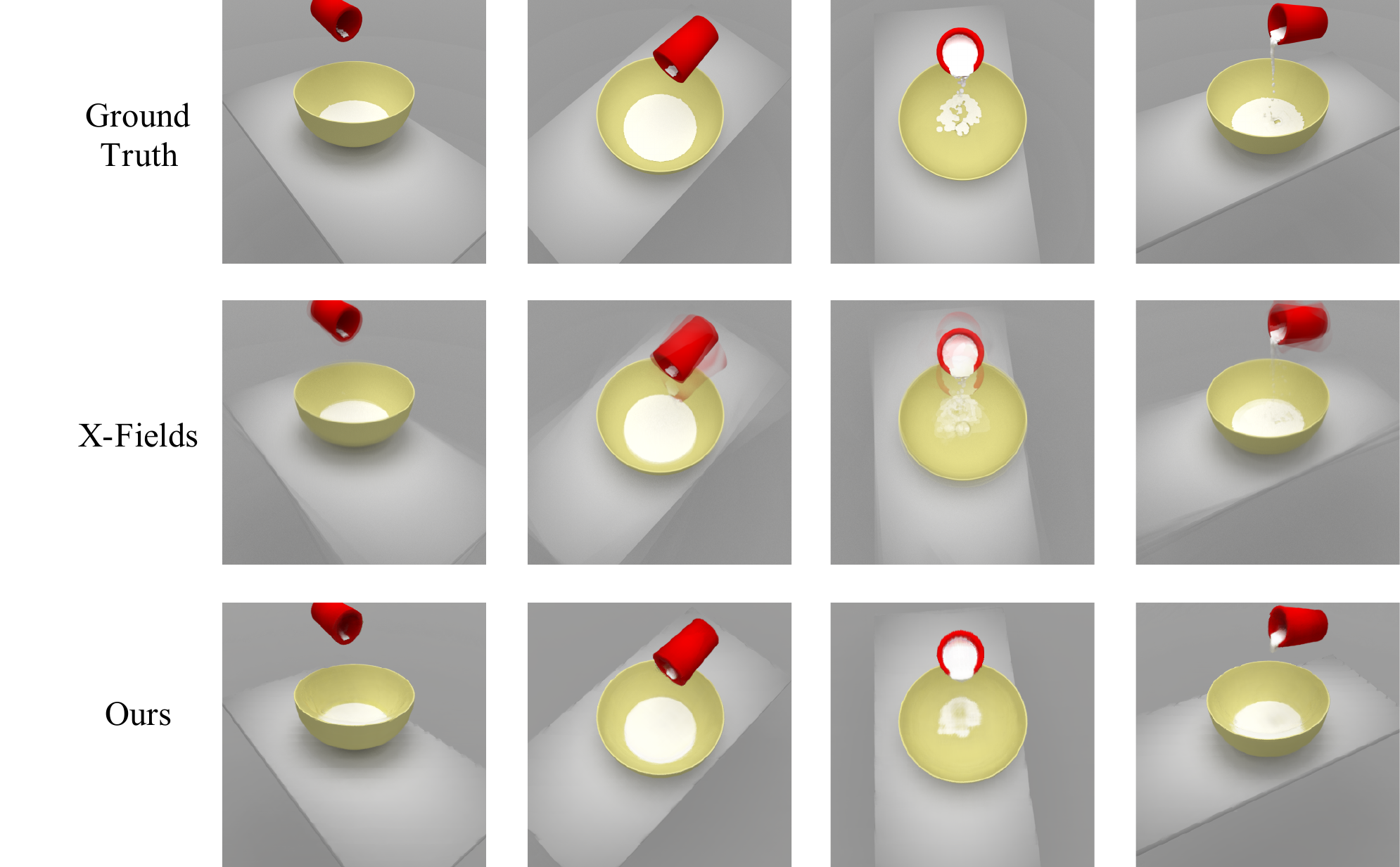}
\caption{Results on Pouring in the Full View setting.
}
\vspace{-10pt}
\label{fig:pouring_gibson_qual_appendix}
\end{figure}

\subparagraph{Full View Synthesis}: We provide full view synthesis results on pouring in \fig{fig:pouring_gibson_qual_appendix}. Compared to X-Fields~\cite{Bemana2020xfields}, we find \model is able to generate more coherent animations of pouring. 

\begin{figure}[h]
\centering
\includegraphics[width=\linewidth]{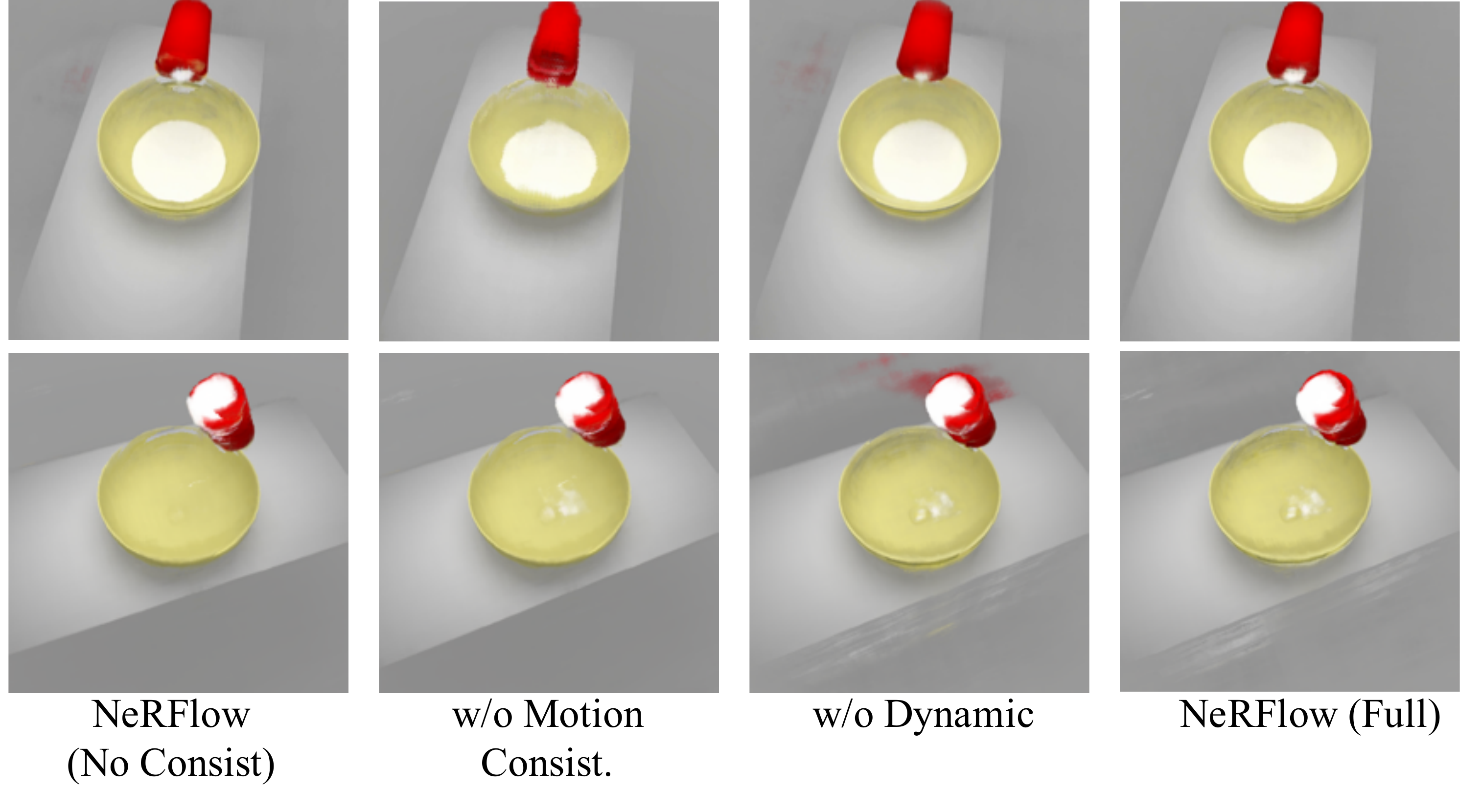}
\caption{Qualitative comparisons of ablations. No consistency and w/o motion consistency have poor bowl structure. Without dynamic modeling, we observe that underlying motion of the cup is blurred across renderings. Finally, \model obtains non-blurry renderings that also capture the underlying bowl structure. 
}
\vspace{-10pt}
\label{fig:ablations_qual}
\end{figure}

\subparagraph{Ablation Visualizations}: We provide visualizations of renderings of each ablation of \model on the stereo capture setting in pouring scene in \fig{fig:ablations_qual}. With either no consistency terms or no motion consistency, the renderings have poor bowl structure. Without dynamic modeling, we observe that the underlying motion of the cup is blurred across renderings. Finally, \model obtains non-blurry renderings that also capture the underlying bowl structure.  

\subparagraph{Supplemental Visualizations}: We recommend readers to examine our attached supplemental video consisting of visualizations of \model. We first show 4D view synthesis on captured monocular videos.
Next, we provide a visualization showing integration of flow across time. We then show examples of free view synthesis as well as view synthesis under stereo and dual camera configurations on Pouring and Gibson scenes. Finally, we show \model applied to video processing tasks of de-noising.

\section{Additional Experimental Details}
\label{sect:training_details}
We train models using the PyTorch framework~\cite{NEURIPS2019_9015}, and utilize the same base architecture as \cite{mildenhall2020nerf} for both radiance and flow functions. Models are trained for 10 hours with only $\mathcal{L}_\text{Render}$, and then trained for another 10 hours with all losses utilizing a single Nvidia 2080 Ti GPU. Models are trained with a learning rate of $0.001$, with a exponential learning rate decay by a factor $0.1$ every 40,000 steps using the Adam optimizer~\cite{Kingma2015Adam}. 

To integrate flow for temporal correspondences, we utilize the Neural ODE library~\cite{chen2018neural}, using the Runga-Kutta solver, with an $\text{RTOL}=10^{-4}$ and $\text{ATOL}=10^{-5}$. While these values are higher than typical values used during Neural ODE training, we found that this is critical for stable flow inference. Smaller values cause much slower training times and lead to less smooth flow fields. 
When training models, we penalize the predictions of $\textbf{c}_{\text{specular}}$ using an $\mathcal{L}_2$ coefficient of $0.1$. We apply coefficients of $0.001$ for $\mathcal{L}_{\text{Corr}}$ and $\mathcal{L}_{\text{Density}}$.  We provide pseudocode for training in Algorithm \ref{alg:pseudocode}

\begin{algorithm}
\small
\begin{algorithmic}
    \STATE \textbf{Input:} Radiance function $R_\theta$, Flow function $F_\theta$, Spatial Correspondences $(\vx_s, t_s), (\vx_g, t_g)$, Camera Rays $C$, RGB values $R$, Timesteps $t$
    \STATE \emph{$\triangleright$ Train NeRFlow Model:}
    \WHILE{not converged}
    \STATE $C_i, R_i, t_i, (\vx_s^i, t_s^i), (\vx_g^i, t_g^i) \leftarrow C, R, t, (\vx_s, t_s), (\vx_g, t_g)$ \quad
    \STATE $\mathcal{L}_\text{Render} = (\text{Render}(R_\theta, t, C_i) - R_i)^2$ \quad \# [41]
    \STATE $\mathcal{L}_\text{Flow} = (\hat{\vx}_g^i -\vx_g^i)^2$ \quad \# $\hat{\vx}_g^i$ computed from flow integration
    \STATE \emph{$\triangleright$ Compute new spatial-temporal correspondance using $F_\theta$:}
    \STATE $(\vx_s', t_s'), (\vx_g', t_g') \leftarrow F_\theta$
    \STATE $\mathcal{L}_\text{Density} =  \|\sigma(\vx_s', t_s') - \sigma(\vx_g', t_g') \|$ \STATE $\mathcal{L}_\text{RGB} =  \|\vc(\vx_s', t_s') - \vc(\vx_g', t_g') \|$
    \STATE \emph{Optimize all losses using Adam:}
    \ENDWHILE
    \STATE \emph{$\triangleright$ Render From NeRFlow Model:}
    \STATE $C \leftarrow V, t$  \quad \# Choose viewpoint $V$ and timestamp $t$
    \STATE \text{Image} = \text{Render}($R_\theta$, $t$, $C$) \quad \# Volumetric rendering [41]
  \end{algorithmic}
 \caption{NeRFlow Training and Sampling Algorithm}
 \label{alg:pseudocode}
 \end{algorithm}

\end{document}